\documentclass[a4paper,fleqn]{cas-dc}
\usepackage{natbib}
\usepackage{subfig}
\usepackage{hyperref}       % hyperlinks
\usepackage{url}            % simple URL typesetting
\usepackage{amsfonts}       % blackboard math symbols
\usepackage{nicefrac}       % compact symbols for 1/2, etc.
\usepackage{microtype}      % microtypography
\usepackage{xcolor}         % colors
\usepackage{amsmath}
\usepackage{amssymb}
\usepackage{graphicx}
\usepackage{tabularx}
\usepackage{tabulary}
\usepackage{multirow}
\usepackage{paralist}
\usepackage{enumitem}
\usepackage{booktabs}
\usepackage{caption}
\usepackage{bm}
\usepackage{amssymb}
\usepackage[switch]{lineno}
\usepackage[ruled,linesnumbered,lined,boxed,commentsnumbered]{algorithm2e}

\newcommand{\tablestyle}[2]{\setlength{\tabcolsep}{#1}\renewcommand{\arraystretch}{#2}\centering\footnotesize}

\def\tsc#1{\csdef{#1}{\textsc{\lowercase{#1}}\xspace}}
\tsc{WGM}
\tsc{QE}
\tsc{EP}
\tsc{PMS}
\tsc{BEC}
\tsc{DE}
\begin{document} %avoid word overflow
\begin{sloppypar}
\let\WriteBookmarks\relax
\def\floatpagepagefraction{1}
\def\textpagefraction{.001}
\shorttitle{ISPRS Journal of Photogrammetry and Remote Sensing}
\shortauthors{H. Zhu, et~al.}

\title [mode = title]{Tiny Object Detection with Single Point Supervision}     

\address[1]{School of Electronic Information, Wuhan University, Wuhan 430072, China}
\address[2]{School of Computer Science, Wuhan University, Wuhan 430072, China}

\cortext[cor1]{Corresponding author}

\author[1]{Haoran Zhu}[style=chinese,orcid=0009-0003-0153-1305]
\ead{zhuhaoran@whu.edu.cn}

\author[1]{Chang Xu}[style=chinese,orcid=0000-0002-3078-0496]
\ead{xuchangeis@whu.edu.cn}

\author[1]{Ruixiang Zhang}[style=chinese,orcid=0000-0003-0704-2484]
\ead{zhangruixiang@whu.edu.cn}

\author[2]{Fang Xu}[style=chinese,orcid=0000-0003-4260-7911]
\ead{xufang@whu.edu.cn}

\author[1]{Wen Yang}[style=chinese,orcid=0000-0002-3263-8768]
\cormark[1]
\ead{yangwen@whu.edu.cn}

\author[1]{Haijian Zhang}[style=chinese,orcid=0000-0001-8314-6563]
\ead{haijian.zhang@whu.edu.cn}

\author[2]{Gui-Song Xia}[style=chinese,orcid=0000-0001-7660-6090]
\ead{guisong.xia@whu.edu.cn}

\begin{abstract}
Tiny objects, with their limited spatial resolution, often resemble point-like distributions. As a result, bounding box prediction using point-level supervision emerges as a natural and cost-effective alternative to traditional box-level supervision.
However, the small scale and lack of distinctive features of tiny objects make point annotations prone to noise, posing significant hurdles for model robustness.
To tackle these challenges, we propose Point Teacher—the first end-to-end point-supervised method for robust tiny object detection in aerial images.
To handle label noise from scale ambiguity and location shifts in point annotations, Point Teacher employs the teacher-student architecture and decouples the learning into a two-phase denoising process. In this framework, the teacher network progressively denoises the pseudo boxes derived from noisy point annotations, guiding the student network's learning. Specifically, in the first phase, random masking of image regions facilitates regression learning, enabling the teacher to transform noisy point annotations into coarse pseudo boxes. In the second phase, these coarse pseudo boxes are refined using dynamic multiple instance learning, which adaptively selects the most reliable instance from dynamically constructed proposal bags around the coarse pseudo boxes.
Extensive experiments on three tiny object datasets (\textit{i.e.}, AI-TOD-v2, SODA-A, and TinyPerson) validate the proposed method's effectiveness and robustness against point location shifts. 
Notably, relying solely on point supervision, our Point Teacher already shows comparable performance with box-supervised learning methods.
Codes and models will be made publicly available.
\end{abstract}


\maketitle
% introduction
\section{Introduction}
Despite the remarkable progress made in tiny object detection recently~\citep{smallod_uav, Dense_TOD, dqdetr}, the success of modern tiny object detectors largely depends on the availability of large-scale high-quality annotations such as TinyPerson~\citep{TinyPerson_2020_WACV}, AI-TOD~\citep{AI-TOD_2020_ICPR, aitodv2_2022_isprs}, and SODA~\citep{soda_2023_pami}. However, obtaining high-quality annotations is particularly challenging for tiny objects, their intrinsic characteristics\textendash occupying few pixels (less than 16 $\times$ 16) and lacking discriminative features\textendash significantly increase the cost and difficulty of box annotation. 
Moreover, the extremely limited pixel footprint of tiny objects leads to highly sparse shape and feature information, making them resemble point-like distributions. This naturally raises an intriguing question: \textit{can we simplify box annotations as point annotations to supervise tiny object detection?}

\begin{figure}[t]
    \centering
    \includegraphics[width=0.99\linewidth]{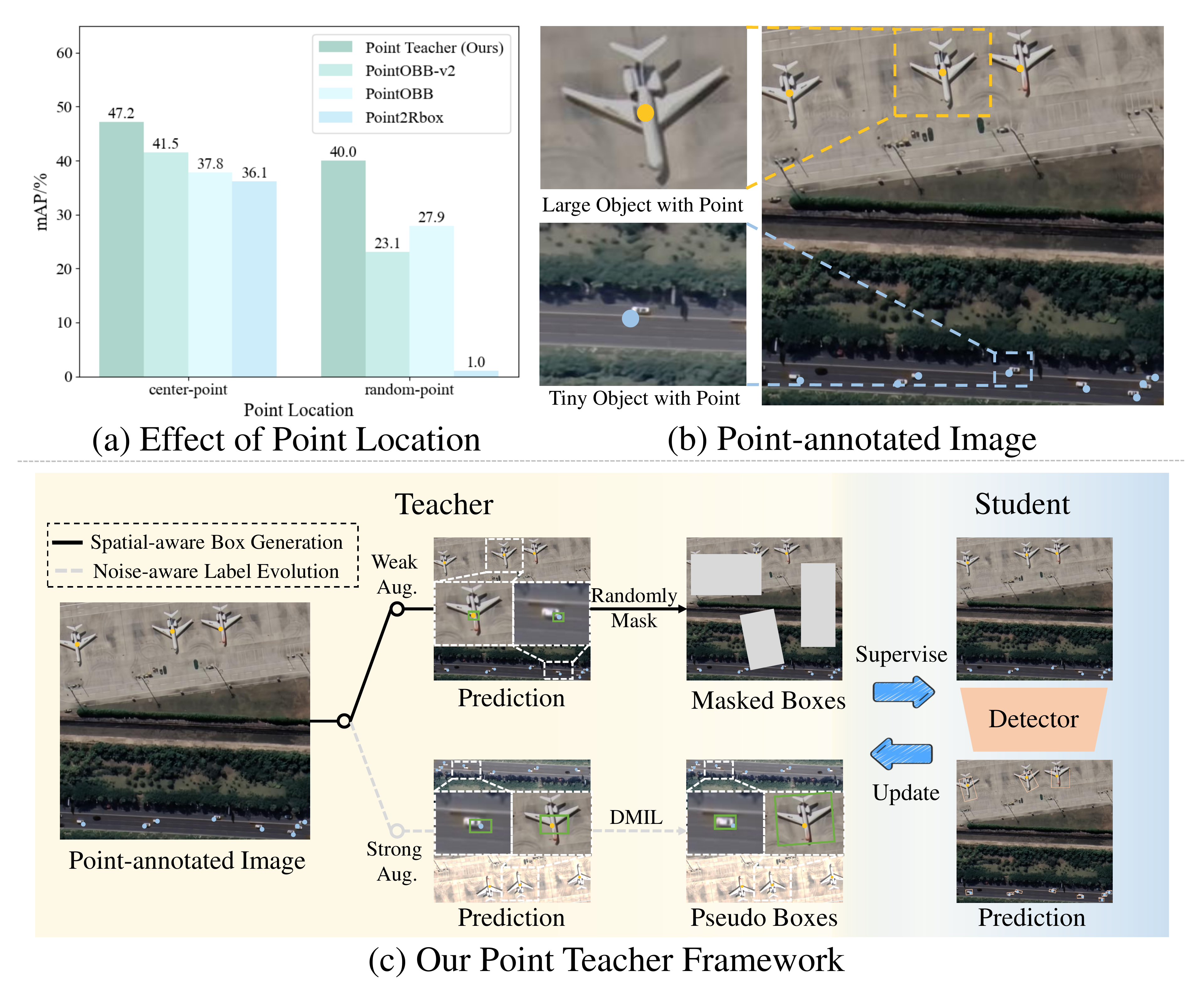}
    \caption{(a) Effect of point location on accuracy: previous methods assume that the point location lies within the center region, and performance significantly degrades when the point location slights shifts around the center. (b) Comparison of point annotations for large and tiny objects: the limited scale and ambiguous boundaries make it challenging to annotate accurately on the main body of the tiny object. (c) An overview of our proposed Point Teacher: we propose a one-step two-phase learning paradigm that is robust to point location, consisting of Spatial-aware Box Generation and Noise-aware Label Evolution.}
    \label{fig:overall}
\end{figure}

This question, although previously explored for generic objects~\citep{p2bnet, plug, pointobb, point2rbox, end2endpsod, polo}, remains a significant challenge for tiny object detection due to the unique issues of small scale and ambiguous boundary.
Specifically, existing point-supervised object detection (PSOD) methods generally impose strong prior assumptions on the point location, such as center-point~\citep{point2rbox}, center-region~\citep{pointobb}, gaussian-region~\citep{p2bnet}, or mask-region~\citep{plug} priors. 
These assumptions, despite being beneficial for optimizing models in the generic scenarios, will not hold anymore for tiny objects, leading to a collapse in performance (Figure~\ref{fig:overall}(a)). 
This collapse can be largely attributed to the noisy distribution of point annotations for tiny objects: the extremely limited scale and ambiguous boundaries not only make it challenging to ensure that point annotations accurately fall on the main body of the object but also render models particularly susceptible to location noise (Figure~\ref{fig:overall}(b)).

To bridge the gap between the challenges posed by noisy point annotations and the need for accurate tiny object detection, we introduce \textbf{Point Teacher} (Figure~\ref{fig:overall}(c))\textemdash a point-supervised method robust to location noise with a denoising-based training approach.
In the previous point-to-box training, the lack of scale information and noisy point annotations significantly reduce the quality of generated pseudo boxes, severely undermining the effectiveness of the supervision signal for box prediction. 
Point Teacher tackles this challenge by converting the PSOD task for tiny objects into a two-phase box generation and denoising process: the first phase converts noisy point annotations to coarse box predictions, and the second phase progressively refines box quality by learning to denoise bounding boxes.
We name each phase as \textbf{Spatial-aware Box Generation} and \textbf{Noise-aware Label Evolution}, respectively.
Despite different phases, the entire model is trained in an end-to-end manner. Remarkably, unlike previous methods that rely on auxiliary knowledge (\textit{e.g.}, synthesizing objects or using SAM-based models) for point-to-box generation, our approach eliminates the usage of auxiliary knowledge with a self-supervised learning strategy that enables the network to directly infer the coarse scale of tiny objects. 

More specifically, in the Spatial-aware Box Generation phase, we randomly mask out parts of the image and enforce the network to predict the scale and location of the masked regions, granting the network an initial sense of spatial awareness (\textit{i.e.}, box regression ability). 
In the subsequent Noise-aware Label Evolution phase, we introduce a Dynamic Multiple Instance Learning (DMIL) module to refine the noisy pseudo boxes generated by the teacher network, providing cleaner supervision for the student network. 
Compared to previous MIL modules, our DMIL dynamically extends object bags and corrects each proposal's locations within the bag. This location adjustment enhances the reliability of bag generation, even when point annotations are noisy.
Additionally, we propose a simple yet effective noise-robust regression loss, termed Jittering IoU Loss, to mitigate overfitting to noisy pseudo boxes. Jittering IoU Loss applies controlled, small perturbations to the regression targets, encouraging the model to learn from multiple nearby target locations. This allows the model to better capture the overall target distribution, avoiding overfitting to specific noisy boxes.

Our Point Teacher can be seamlessly integrated into various detector architectures, supporting both horizontal bounding box (HBB) and oriented bounding box (OBB) tasks. Comprehensive experiments conducted on tiny object datasets (\textit{i.e.}, AI-TOD-v2, TinyPerson, and SODA-A) demonstrate the robustness and effectiveness of the proposed method.
The main contributions of this paper are three-fold: 

\begin{itemize}
    \item We propose Point Teacher, the first end-to-end point-supervised framework for tiny object detection, specifically designed to address the challenge of achieving accurate detection under noisy point annotations.
    \item Our Point Teacher decouples the learning process into a two-phase denoising learning paradigm, comprising the Spatial-aware Box Generation phase and Noise-aware Label Evolution phase, to ensure robust performance under noisy point supervision.
    \item We demonstrate that our Point Teacher is highly generalizable to off-the-shelf object detectors and supports both horizontal bounding box (HBB) and oriented bounding box (OBB) tasks, achieving state-of-the-art results on point-based tiny object datasets under both center-based and noisy point annotation.
\end{itemize}

The rest of this paper is organized as follows. In Section~\ref{relatedwork}, we briefly survey related works. In Section~\ref{sec:Methodology}, we introduce the details of Point Teacher. Then, we validate the robustness and effectiveness of the proposed method in Section~\ref{experiments} and provide a detailed discussion in Section~\ref{discussion}. Finally, we conclude this paper in Section~\ref{conclusion}.

\begin{figure*}[t]
    \centering
    \includegraphics[width=1.0\linewidth]{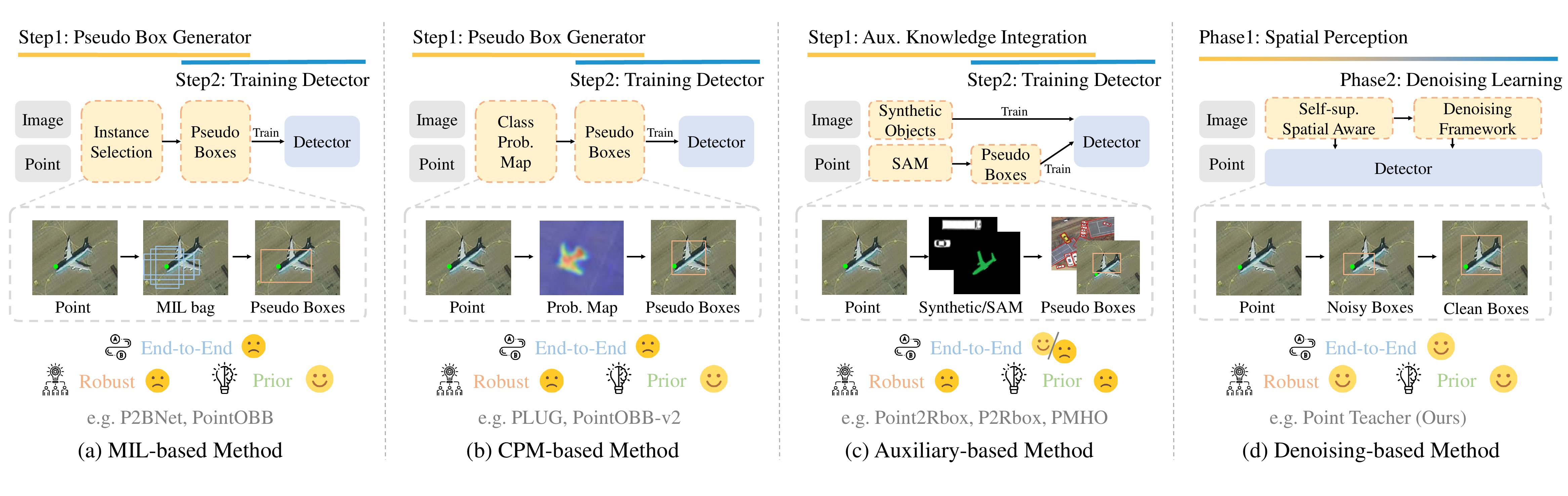}
    \caption{A comparison with existing point-supervised object detection methods, including (a) MIL-based methods; (b) CPM-based methods; (c) Auxiliary-based methods; (d) Denoising-based methods. (a), (b), and (c) paradigms adopt a two-step, non-end-to-end training process. (d) paradigm adopts a one-step, two-phase end-to-end training process. SAM denotes Segment Anything Model.}
    \label{fig:compare_PSOD}
\end{figure*}

% related work
\section{Related Work}
\label{relatedwork}
\subsection{Tiny Object Detection}
\label{relatedwork:FullTOD}
The extremely limited pixel count in tiny objects poses significant challenges for generic object detectors, leading to a surge in specialized research targeting this issue. In short, we can distinguish methods designed for tiny object detection as follows. 

(a) \textbf{Multi-scale image and feature representations.} 
At the image level, techniques such as SNIP~\citep{SNIP_2018_CVPR} and SNIPer~\citep{SNIPER_2018_NIPS} normalize object scales within a specific range to achieve scale-invariant detection. At the feature level, methods like Feature Pyramid Networks (FPN) have become foundational for multi-scale detection strategies~\citep{FPN_2017_CVPR, hynet_2021_isprs, msod_2018_isprs}, with advancements including PANet~\citep{PANet_2018_CVPR}, Recursive-FPN~\citep{Efficientdet_2020_CVPR}, BiFPN~\citep{DetectoRS_2020_CVPR}, TridentNet~\citep{Trident-Net_2019_ICCV}, and Denoising-FPN~\citep{DFPN}, which refine these approaches. Recent works by Wang et al.\citep{wang2024small}, Xiao et al.\citep{Fusion}, and Peng et al.~\citep{lgffyolo} have introduced novel feature fusion strategies that enhance the representation of tiny objects by effectively integrating both global and local features.
(b) \textbf{Super resolution.} 
In addition to multi-scale techniques, super-resolution-based detectors have emerged as a powerful approach for enhancing the feature representation by effectively reconstructing high-resolution features from limited pixel information~\citep{SOD-MTGAN_2018_ECCV, PGAN_2017_CVPR, Better_to_Follow_2019_ICCV, msod_2018_isprs, Super-Resolution_jstars}. These methods leverage advancements in generative models and image enhancement to amplify subtle details, thereby improving detection accuracy for tiny objects.
(c) \textbf{Learning strategies.}
Recent advances have also shed new light on the learning strategies for tiny object detection (TOD). Notably, recognizing the vulnerability of Intersection over Union (IoU) to box offsets, several works have introduced new metrics for more accurate assignment~\citep{aitodv2_2022_isprs, dotd_2021_cvprw, bdistance_grsl_2022, simd}. Furthermore, more recent approaches~\citep{rfla, dcfl_cvpr_2023} (\textit{e.g.,} RFLA) have developed scale-balanced assignment strategies to provide more effective supervision for tiny objects.

Previous works have made significant strides in tiny object detection under the assumption of a fully labeled training set. However, in real-world scenarios, obtaining fully annotated datasets is prohibitively costly. This work, instead, seeks to achieve robust and efficient learning under point supervision.

\subsection{Point-supervised  Object Detection}
\label{relatedwork:PointTOD}
Training with point annotations has garnered considerable attention due to its low annotation cost. To achieve bounding box prediction with only point supervision, a widely adopted approach is the two-step point-to-box conversion. The first step involves training a model to generate pseudo boxes from point annotations, while the second step utilizes the generated pseudo boxes to train the object detector. Based on the way of generating pseudo boxes, existing methods can be categorized as follows (Figure~\ref{fig:compare_PSOD}).

(a) \textbf{MIL-based methods.} 
MIL-based methods typically begin by training a Multiple Instance Learning (MIL) model to generate pseudo boxes, which are then used to train the detector. 
Papadopoulos et al.~\citep{clicksupervision} have proposed center-click annotation as a replacement for box annotation, using MIL to refine the localization process. UFO2~\citep{ufo2} introduces a unified weakly supervised detection framework, leveraging MIL to learn and localize targets from various types of annotations such as tags, points, scribbles, or boxes. Omni-DETR~\citep{omnidetr} extends UFO2 by supporting more forms of mixed annotations, leading to improved detection accuracy. However, these MIL methods are based on OTSP approaches and are not specifically designed for point supervision tasks. 
P2BNet~\citep{p2bnet} first introduces an improved MIL framework tailored for point supervision, significantly improving the quality of pseudo-box generation. 
PointOBB~\citep{pointobb} builds upon P2BNet by incorporating a self-supervised loss to learn angle and scale information, applying it to oriented object detection. 
Zhang et al.~\citep{onepointallyouneed} extend these methods to a sparse point annotation setting, substantially reducing the annotation cost.
(b) \textbf{CPM-based methods.} 
CPM-based methods train a classification head to produce a Class Probability Map (CPM), which is subsequently used to generate pseudo boxes for training the detector. PLUG~\citep{plug} refines the point-to-box process by introducing a point-mask-box framework, where the CPM is used to generate masks that facilitate the generation of pseudo boxes. PointOBB-v2~\citep{pointobbv2} enhances this process by proposing a non-uniform positive and negative sampling strategy for training the CPM, thereby achieving more accurate mask generation.
(c) \textbf{Auxiliary-based methods.}
Auxiliary-based methods consist of two subcategories distinguished by the source of auxiliary knowledge: synthetic-based and SAM-based methods. Synthetic-based methods manually synthesize objects or patterns as pseudo labels for end-to-end training, while SAM-based methods refine the detection process using masks generated by the Segment Anything Model (SAM)~\citep{sam}.
Point2Rbox~\citep{point2rbox}, as a synthetic-based method, introduces synthetic knowledge and constructing synthetic objects to learn regression capabilities, enabling end-to-end oriented object detection.
P2RBox~\citep{p2rbox} and PMHO~\citep{pmho} adopt the point-mask-box paradigm and integrate the SAM~\citep{sam} model, resulting in a significant performance boost for the network.

While these methods have advanced point-supervised object detection, they primarily assume that point locations are centered or within central regions. This assumption is too rigorous for tiny objects with minimal pixel occupancy, as even slight positional shifts can lead to a significant drop in accuracy due to the sensitivity of tiny objects to location. 
Additionally, the aforementioned methods adopt a two-step, non-end-to-end paradigm, which typically requires more time to implement the point-to-box detection training process.
Thus, in this paper, we focus on the impact of point location on network performance and propose Point Teacher, a robust, end-to-end denoising-based PSOD method.

\begin{figure*}[t]
    \centering
    \includegraphics[width=0.99\linewidth]{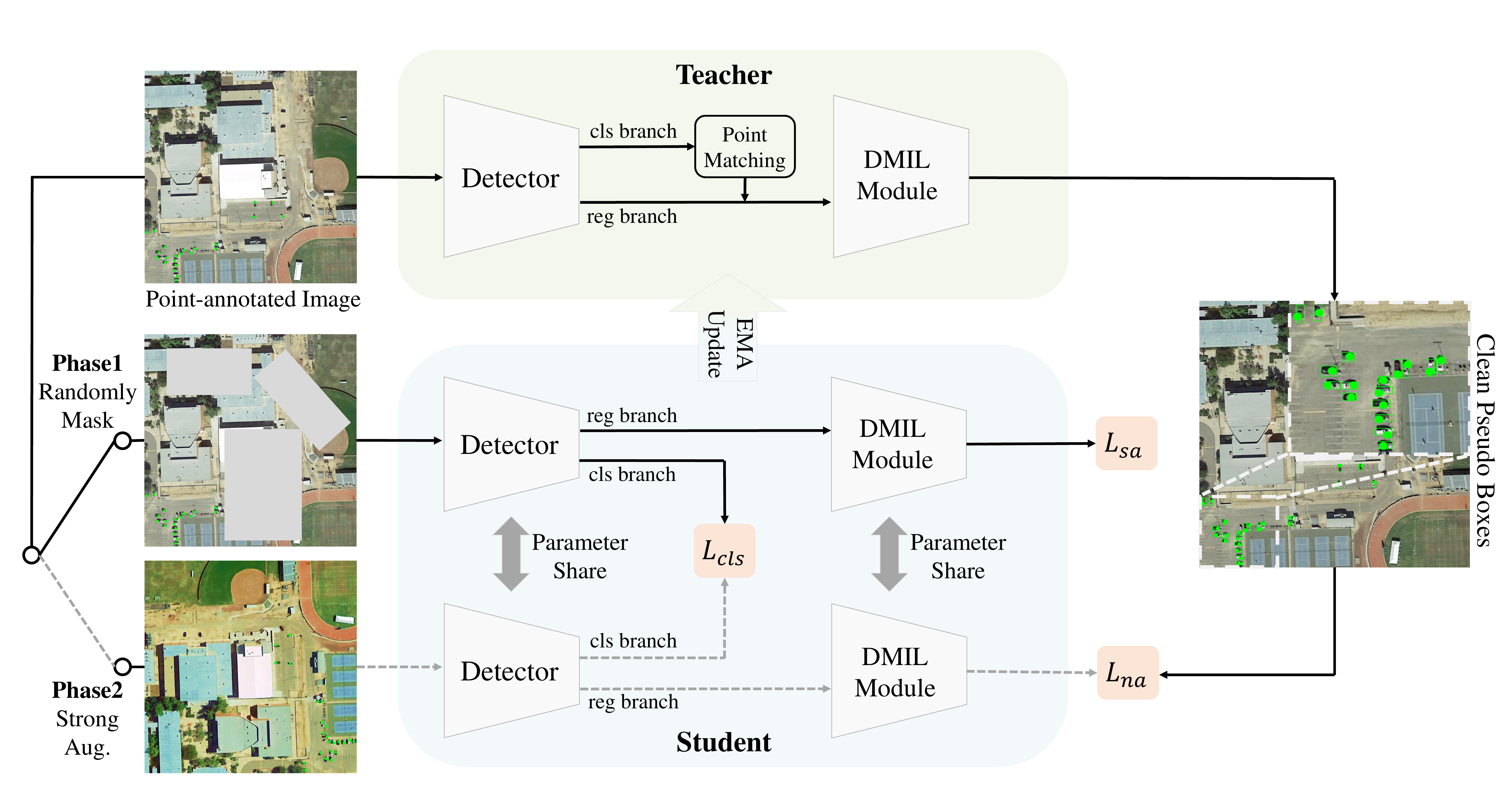}
    \caption{The framework of Point Teacher. The training process of Point Teacher consists of two phases: Spatial-aware Box Generation (phase1) and Noise-aware Label Evolution (phase2). During the Spatial-aware Box Generation phase, the masked image is used to train both the regression branch and the DMIL module, enabling the model to develop spatial awareness. In the Noise-aware Label Evolution phase, the teacher network, in conjunction with the DMIL module, generates clean pseudo boxes to supervise the student network for end-to-end learning. The classification learning is integrated throughout the phases.}
    \label{fig:overall_framework}
\end{figure*}

% method
\section{Methodology}
\label{sec:Methodology}
\subsection{Overall Framework}
\label{overall framework}
% This work aims to train a point-supervised tiny object detector in an end-to-end manner. Unlike previous non end-to-end approaches~\citep{p2bnet, point2rbox, pointobb, pointobbv2}, which first train a pseudo-boxes generator and then a detector, we propose a two-phase denoising learning paradigm: Spatial-aware Box Generation and Noise-aware Label Evolution, as illustrated in Figure~\ref{fig:overall_framework}.
This work proposes an end-to-end point-supervised tiny object detector. Unlike previous two-step methods~\citep{p2bnet, point2rbox, pointobb, pointobbv2}, which first train a pseudo-box generator and then a detector, we introduce an end-to-end denoising-based method consisting of the Spatial-aware Box Generation phase and Noise-aware Label Evolution phase, as shown in Figure~\ref{fig:overall_framework}. Both phases are integrated into a unified pipeline, enabling direct optimization from point annotations to the final detection output. In the Spatial-aware Box Generation phase, the network is trained to develop spatial awareness and learn the mapping from points to coarse pseudo boxes. 
In the subsequent Noise-aware Label Evolution phase, the network undergoes denoising learning to refine the coarse pseudo boxes into precise pseudo boxes.
Additionally, as the annotated point provides both class information and rough positional cues, object classification is performed throughout the entire process. The overall loss function can be summarized as:
\begin{equation}
    L = L_{cls} + L_{sa} + L_{na}, 
    \label{con:overall_loss}
\end{equation}
where $L_{cls}$ represents the classification loss from the detection head. $L_{sa}$ and $L_{na}$ denote the loss during the Spatial-aware Box Generation phase and Noise-aware Label Evolution phase, respectively.
During inference, only the detector is utilized. 

Additionally, since Point Teacher is orientation-agnostic, it is applicable to both Horizontal Bounding Box (HBB) and Oriented Bounding Box (OBB) detection tasks. In the final part of this section, we will demonstrate how to deploy the method on HBB detectors. For OBB detectors, the only required adjustment is the incorporation of the angle parameter $\theta$.

\begin{figure*}[t]
    \centering
\includegraphics[width=0.99\linewidth]{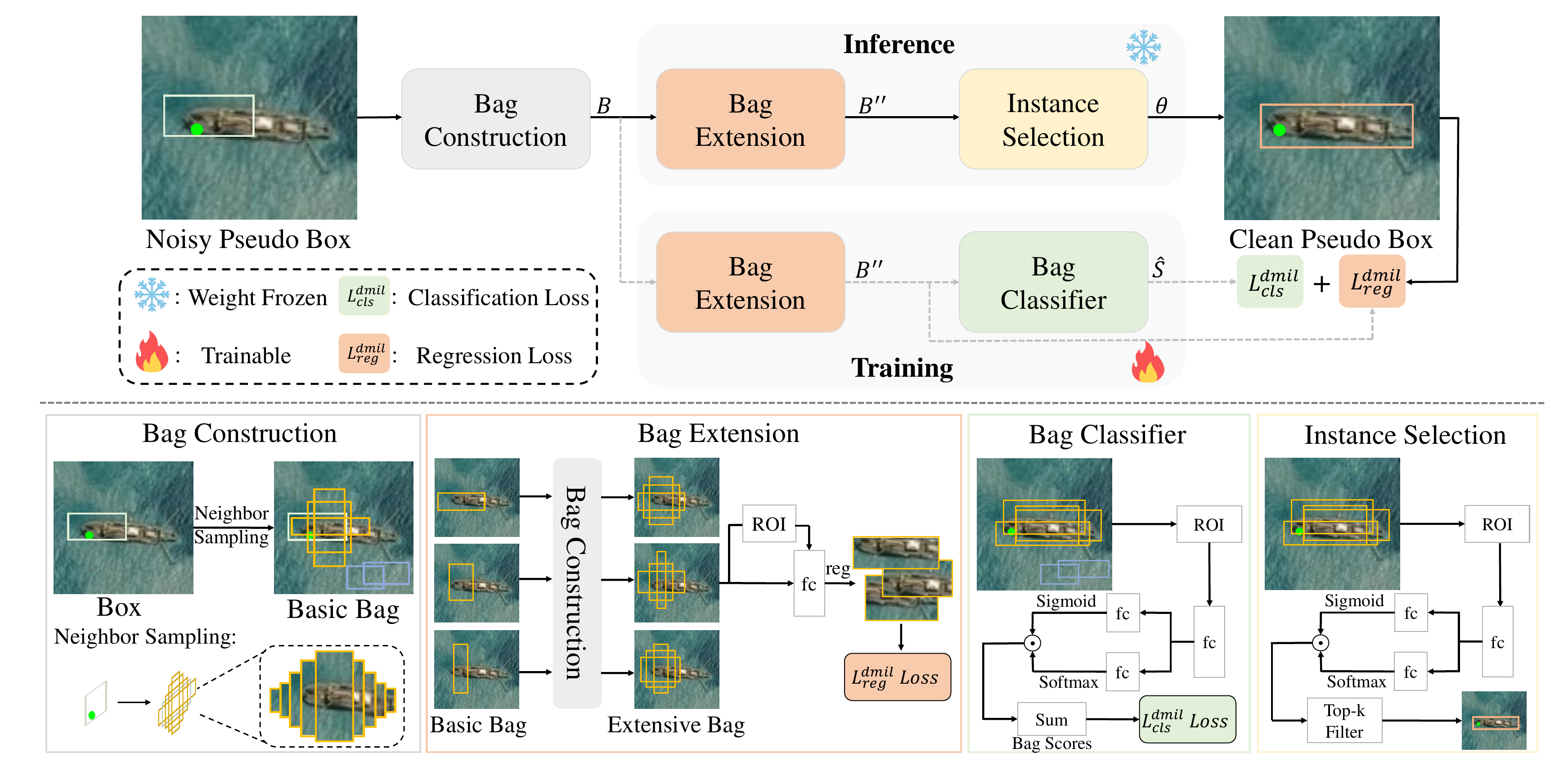}
    \caption{The workflow of Dynamic Multiple Instance Learning Module (DMIL). DMIL comprises four stages: Bag Construction, Bag Extension, Bag Classifier, and Instance Selection. The Bag Construction and Bag Extension stages ensure the creation of high-quality bags. The Bag Classifier is used to train the classification and discrimination capabilities of the bags. Finally, Instance Selection uses the scores from the Bag Classifier to merge instances, generating pseudo boxes.}
    \label{fig:dmil}
\end{figure*}

\subsection{Spatial-aware Box Generation}
\label{Spatial-aware Box Generation}
The point-based annotation lacks both scale and precise location information of objects, making it impractical to regress bounding box solely from point supervision. 
Inspired by DINOv2's self-supervised approach for learning robust visual features, which masks random image patches to ensure feature consistency~\citep{dinov2}, we adapt this strategy for the box regression task to help the network develop spatial awareness.
Specifically, we randomly mask certain regions of the image: $R$ ($\{cx, cy, w, h\}$ for HBB task and $\{cx, cy, w, h, \theta\}$ for OBB task). The regression head is then tasked with predicting the scale and position of the masked region.
The supervision for this task is provided by the loss term $L_{reg}^{sa}$, which is computed for each masked region $R$.
\begin{equation}
    L_{reg}^{sa} = L_{bbox}(P, R),
    \label{con:per_reg}
\end{equation}
where $P$ denotes the network's prediction, with $P$ representing $\{cx, cy, w, h\}$ for the HBB task and $\{cx, cy, w, h, \theta\}$ for the OBB task. The term $L_{bbox}$ represents the regression loss, such as Smooth L1 Loss or IoU-based Loss. In this work, we use the Jittering IoU Loss, which will be described in detail in the following section~\ref{Noise-aware Label Evolution}.

% Mask region regression enhances the model's spatial awareness through a regression-driven guidance mechanism. However, MIL-based methods, which rely on fixed jittering and classification for top-$k$ proposal selection, do not offer such spatial guidance~\citep{SSDdet}.
Mask region regression enhances the model's spatial awareness by leveraging a regression-driven guidance mechanism. However, MIL-based methods construct bags by applying fixed-scale jittering around coarse pseudo boxes, which lacks the flexibility to dynamically adjust bag construction based on spatial awareness~\citep{SSDdet}.
To address this limitation, we introduce a Dynamic Multiple Instance Learning (DMIL) module. As illustrated in Figure~\ref{fig:dmil}, DMIL comprises four key processes: \textbf{Bag Construction}, \textbf{Bag Extension}, \textbf{Bag Classifier}, and \textbf{Instance Selection}. Our DMIL emphasizes extending the constructed bags and introducing a regression branch to refine proposals, enabling improved spatial awareness and more accurate pseudo-box generation.
For clarity, we illustrate the DMIL framework using the HBB task as an example. When extending to the OBB task, it suffices to incorporate the angle parameter $\theta$ into all bounding boxes and replace the RoI extractor~\citep{faster-rcnn} with a Rotated RoI extractor~\citep{RoI-Transformer_2019_CVPR}.

Given an image containing $M$ coarse pseudo boxes $Z \in \mathbb{R}^{M \times 4}$ (generated by the teacher network) and $N$ masked regions $R \in \mathbb{R}^{N \times 4}$. we first perform \textbf{Bag Construction} for each box in $Z$ and $R$. This process generates pseudo-box proposal bags $B_{Z} \in \mathbb{R}^{M \times U_1 \times 4}$, mask proposal bags $B_{R} \in \mathbb{R}^{N \times U_1 \times 4}$, and negative proposals $B_{neg} \in \mathbb{R}^{U_{neg} \times 4}$, where $U_1$ and $U_{neg}$ represent the number of proposals in each bag and the number of negative proposals in an image. 
The bag construction process follows the neighbor sampling strategy used in P2BNet~\citep{p2bnet}. 

Since all the bags are sampled near the coarse pseudo boxes, they may be inaccurate. Therefore, in \textbf{Bag Extension} stage, we extend and refine the bags to ensure that more precise proposals can be sampled. 
For each proposal in bag $B_Z$ and $B_R$, we perform neighbor sampling to expand it into an augmented set, $B_Z^{'} \in \mathbb{R}^{M \times U_1 \times U_2 \times 4}$ and $B_R^{'} \in \mathbb{R}^{N \times U_1 \times U_2 \times 4}$, where $U_2$ denotes the number of newly generated proposals during the bag extension process. Then, we refine the proposals in $B_Z^{'}$ and $B_R^{'}$ to obtain more accurate proposal bags. 
Specifically, using $7\times7$ RoIAlign and two fully connected (fc) layers, the features of the proposals in $B_Z^{'}$ and $B_R^{'}$ are extracted, denoted as $F_Z^{'} \in \mathbb{R}^{M \times U_1 \times U_2 \times D}$ and $F_R^{'} \in \mathbb{R}^{N \times U_1 \times U_2 \times D}$. The regressor then takes the proposal bag and features as input and outputs a refined bag $B_Z^{''} \in \mathbb{R}^{M \times U_1 \times U_2 \times 4}$ and $B_R^{''} \in \mathbb{R}^{N \times U_1 \times U_2 \times 4}$. 
Note that at this phase (\textit{i.e.,} Spatial-aware Box Generation phase), we possess the reliable localization information of masked regions $R$. Thus, we can leverage them to supervise the training of the regressor. Since the supervision signals within the same bag are consistent, we replicate $R$ by $U_1\times U_2$ times to obtain $R' \in \mathbb{R}^{N \times U_1 \times U_2 \times 4}$. The corresponding loss is denoted as:
\begin{equation}
    L_{reg}^{dmil} = L_{bbox}(B_R^{''}, R').
    \label{con:dmil_reg}
\end{equation}

After constructing and extending the bags, it is essential to equip DMIL with the ability to select high-quality proposals from each bag. Thus, in the \textbf{Bag Classifier} stage, we focus on training DMIL's classifier and instance selector.
It is worth noting that since the $R$ lacks class information, we do not apply the Bag Classifier and Instance Selection operations to $B_{R}^{''}$. 
Consequently, we pass $B_{Z}^{''}$ through RoIAlign and two fully connected (fc) layers (without sharing weights with the regressor's fc) to generate features $F_{Z}^{''} \in \mathbb{R}^{M \times U_1 \times U_2 \times D}$. We then apply the classification branch $f_{cls}$ to $F_{Z}^{''}$, yielding $O^{cls} \in \mathbb{R}^{M \times U_1 \times U_2 \times C}$, which is then passed through the $\textit{sigmoid}$ function over the classification dimension to obtain the score $S^{cls}$, where $C$ represents the number of category. Meanwhile, we apply the instance branch $f_{ins}$ to $F_{Z}^{''}$, yielding $O^{ins} \in \mathbb{R}^{M \times U_1 \times U_2 \times C}$, and $S^{ins}$ is obtained through $\textit{softmax}$ function over $U_2$ proposals. 
During training, the score $S$ for each proposal is computed as the Hadamard product of $S^{cls}$ and $S^{ins}$. Subsequently, the scores of all proposals within the bag are summed to obtain the final score for each bag, $\hat{S} \in \mathbb{R}^{M \times U_1 \times C}$.
\begin{equation}
    \begin{cases}
    O^{cls}=f_{cls}(F_{Z}^{''}); \quad
    S_{i,j,k,c}^{cls} = 1 / (1 + e^{-O_{i,j,k,c}^{cls}})\\
    O^{ins}=f_{ins}(F_{Z}^{''}); \quad 
    S_{i,j,k,c}^{ins} = e^{O^{ins}_{i,j,k,c}} / \sum_{x=1}^{U_2} e^{O^{ins}_{i,j,x,c}} \\
    S = S^{cls} \odot S^{ins}; \quad
    \hat{S}_{i,j,c} = \sum_{k=1}^{U_2} S_{i,j,k,c} ,
    \end{cases}
    \label{con:S}
\end{equation}
where $S_{i,j,k,c}$ represents the element value at the $(i, j, k,c)$ index. The $\textit{sigmoid}$ activation is applied to each position $(i, j, k,c)$ of $S^{cls}$ along the class dimension $C$, while the $\textit{softmax}$ activation is applied to each position $(i, j, k,c)$ of $S^{ins}$ along the proposal dimension $U_2$.

The supervision for training DMIL's classifier and instance selector is provided by the loss term $L_{cls}^{dmil}$, which is computed for each bag $B_{Z}^{''}$ and the negative proposals $B_{neg}$. Note that negative proposals do not form bags. Thus,  they only have classification scores $\hat{S}^{neg} \in \mathbb{R}^{U_{neg} \times C}$ without instance scores. The loss $L_{cls}^{dmil}$ is defined as follows:
\begin{equation}
    L_{cls}^{dmil} = L_{cls}(\hat{S}, \xi^{pos}) + L_{cls}(\hat{S}^{neg}, \xi^{neg}),
    \label{con:dmil_cls}
\end{equation}
where $\xi^{pos} \in {\{ 0,1\}}^{M \times U_1 \times C}$, $\xi^{neg} \in {\{ 0\}}^{U_{neg} \times C}$. 
The term $L_{cls}$ represents the classification loss. In this work, we use the Focal Loss~\citep{Focal-Loss_2017_ICCV}.

During the \textbf{Instance Selection} stage, classification and instance scores are utilized to select the top-$k$ scoring proposals from each bag as pseudo boxes. Specifically, we first reshape $S$ and $B_Z^{''}$ into $\mathbb{R}^{M \times (U_1\cdot U_2) \times C}$ and  $\mathbb{R}^{M \times (U_1\cdot U_2) \times 4}$. Then, for each coarse pseudo box $Z_j$, we filter the extension bag to select the top-$k$ most accurate proposals, merging them with the coarse pseudo boxes $Z_j$ to generate precise pseudo boxes $\Theta_j$.
\begin{equation}
    \Theta_j = \beta \cdot Z_j + (1-\beta) \cdot \sum_{i=1}^{k} S_{jic} \cdot B''_{ji}, \quad where \sum_{i=1}^{k} S_{jic}=1,
    \label{con:merge}
\end{equation}
$\beta$ represents the weighted fusion coefficient, a hyperparameter between 0 and 1.

In summary, the loss for the Spatial-aware Box Generation phase can be formulated as follows:
\begin{equation}
    L_{sa} = L_{reg}^{sa} + \alpha_{1} \cdot L_{reg}^{dmil} + \alpha_{2} \cdot L_{cls}^{dmil},
    \label{con:preliminaryleanring}
\end{equation}
where $\alpha_{1}$ and $\alpha_{2}$ are set as 0.01 and 0.25 respectively.

\subsection{Noise-aware Label Evolution}
\label{Noise-aware Label Evolution}
After the Spatial-aware Box Generation phase, the network achieves coarse spatial awareness, allowing the teacher network to predict coarse pseudo boxes. In the Noise-aware Label Evolution phase, we further refine the coarse pseudo boxes and perform denoising training.

Following the typical teacher-student architecture~\citep{softteacher}, we use the pseudo boxes generated by the teacher network to supervise the student network for high-quality training. However, our approach differs from previous methods in two key aspects: 
(1) \textbf{Point Matching}: Unlike previous methods like Soft Teacher~\citep{softteacher}, which determine pseudo boxes solely based on a classification score threshold ($\ge$ 0.9), our approach additionally leverages the positional information from point annotations to better guide pseudo box generation.
(2) \textbf{Box Refinement}: The pseudo boxes produced by the teacher network in the PSOD task are more coarse and unsuitable for direct supervision of the student network. Thus, we leverage DMIL and Jittering IoU Loss to refine these pseudo boxes, enabling more precise box generation and robust learning for the regression branch.
Note that, in the previous Spatial-aware Box Generation phase, we solely train DMIL to enhance spatial awareness without refining pseudo boxes. In this phase, however, we will further apply DMIL to refine the pseudo boxes.

\textbf{Point Matching}: Instead of using classification scores alone to determine pseudo boxes, our approach benefits from the guidance provided by the rough positional information of point annotations. To achieve the best matching between the annotated points and the predicted boxes, we propose a two-stage Top-$K$ point matching method. Specifically, we first filter the Top-$K_1$ candidate boxes based on their $L_1$ distance to the annotated points. Next, from these $K_1$ candidates, we select the Top-$K_2$ boxes using a cost matrix, which helps merge and generate pseudo boxes. The cost matrix consists of two main components: classification cost and spatial cost:
\begin{equation}
    \text{Cost}(i,j) = \underbrace{(L_{cls}(s_{ji}, c_j))}_{\textbf{classification\  cost}} + \underbrace{(1 - \mathbb{1}[p_j \  in \  b_{ji}])}_{\textbf{spatial\  cost}},
    \label{con:costmatrix}
\end{equation}
where $j$ is the index of the annotated point, and $i$ is the index of the predicted box. $p_j$ and $c_j$ represent the annotated point and its category. $s_{ji}$ and $b_{ji}$ denote the classification scores and prediction boxes belonging to annotated point $p_j$. The term $(1 - \mathbb{1}[p_j \  in \  b_{ji}])$ indicates that the cost is 0 when the annotated point lies within the predicted box; otherwise, the cost is 1.

Finally, we perform a box fusion based on the classification scores $s$ to generate the coarse pseudo boxes $\Theta_j$. The fusion formula is as follows:
\begin{equation}
    \Theta_j = \sum_{k=1}^{K_2} s_{jk} \cdot b_{jk}, \quad where \sum_{k=1}^{K_2} s_{jk} = 1.
    \label{con:coarsefusionboxes}
\end{equation}

\textbf{Box Refinement}: 
After obtaining the coarse pseudo boxes, the randomness of the annotated point leads to pseudo boxes with potential offsets and scale variations. To provide the student network with high-quality pseudo boxes, we first perform Bag Construction and Bag Extension for $\Theta_j$, generating the candidate proposals Bag $B_j^{''}$ and proposal scores $S_j$. These are then passed to the Bag Classifier and Instance Selection stages, where the Top-$K_3$ most accurate proposals are chosen to generate the refined pseudo boxes $\Theta_j^{'}$. The fusion method is as follows:
\begin{equation}
    \Theta_j^{'} = \beta \cdot \Theta_j + (1-\beta) \cdot \sum_{k=1}^{K_3} S_{jk} \cdot B_{jk}^{''}, \quad where \sum_{k=1}^{K_3} S_{jk} = 1.
    \label{con:refinedfusionboxes}
\end{equation}

During training, we use the generated pseudo boxes $\Theta'$ to supervise both the regression branch of the detection head and the regression branch of DMIL. Additionally, the classifier in DMIL remains unchanged.
The overall loss function is:
\begin{equation}
    \begin{cases}
        L_{reg}^{na} = L_{bbox}(P, \Theta')\\
        L_{reg}^{dmil} = L_{bbox}(B'', \Theta') \\
        L_{na} = L_{reg}^{na} + \alpha_1 \cdot L_{reg}^{dmil} + \alpha_2 \cdot L_{cls}^{dmil}.
    \end{cases}
    \label{con:dn_loss}
\end{equation}

To further enhance the network's robustness and resistance to noisy pseudo boxes, we propose \textbf{Jittering IoU Loss}, a simple yet effective auxiliary regression loss. The loss encourages the model to learn across different, yet nearby target positions, enabling it to better capture the overall object distribution and avoid overfitting to specific noisy pseudo boxes~\citep{positive_noise}. Specifically, for HBB task, given a predicted box $A=(\textit{cx},\textit{cy},\textit{w},\textit{h})$ and a regression target $B_{gt}=(\textit{cx}_{gt},\textit{cy}_{gt},\textit{w}_{gt},\textit{h}_{gt})$, we first expand and shrink $B_{gt}$ by a certain ratio $r$ to create perturbed versions $B'_{gt}$:
\begin{equation}
    B'_{gt}=
    \begin{cases}
        (\textit{cx}_{gt},\ \textit{cy}_{gt},\ \textit{w}_{gt},\ \textit{h}_{gt}) \\
        (\textit{cx}_{gt},\ \textit{cy}_{gt},\ \textit{w}_{gt}-\textit{w}_{gt} \cdot r,\ \textit{h}_{gt}-\textit{h}_{gt} \cdot r) \\
        (\textit{cx}_{gt},\ \textit{cy}_{gt},\ \textit{w}_{gt}-\textit{w}_{gt} \cdot r,\ \textit{h}_{gt}+\textit{h}_{gt} \cdot r) \\
        \qquad \qquad \qquad ...\\
        (\textit{cx}_{gt},\ \textit{cy}_{gt},\ \textit{w}_{gt}+\textit{w}_{gt} \cdot r,\ \textit{h}_{gt}+\textit{h}_{gt} \cdot r).
    \end{cases}
    \label{con:shakingbag}
\end{equation}

For the OBB task, we keep the angle parameter unchanged and only perturb the parameters $(\textit{cx},\textit{cy},\textit{w},\textit{h})$.
The final regression loss is composed of the base loss combined with the minimum loss derived from the perturbed targets.
\begin{equation}
    L_{bbox} = L_{IoU}(A, B_{gt}) + \min_{i} L_{IoU}(A, B'_{gt}[i]). 
\end{equation}

\subsection{Detector Integration}
Our method is general and is not limited to specific object detectors.
However, due to the lack of scale information in point annotations, scale-aware components like FPN~\citep{FPN} and Label Assignment in existing detectors such as FCOS~\citep{fcos} and Faster R-CNN~\citep{faster-rcnn} cannot be directly used. To address this issue, we replace the FPN and Label Assignment with the proposed \textbf{Top-down FPN Aggregation} and \textbf{Scale-invariant Label Assignment} in our method.

\textbf{Top-down FPN Aggregation}: Each layer of the FPN has feature points with varying receptive fields~\citep{fpnagg}, typically used for detecting objects of different sizes, from small to large, across layers $P_3$ to $P_7$. For tiny objects, the features are mainly allocated to the $P_3$ layer. To avoid scale confusion while still incorporating high-level semantic information, we propose a simple and effective Top-down FPN Aggregation strategy. Specifically, we use $1 \times 1$ convolution ($\textit{Conv}$) and Upsampling ($\textit{Up}$) operations to aggregate features from layers $P_3$ to $P_7$ into a single output layer $M$, as shown below:
\begin{equation}
    \begin{cases}
        P_{i-1} = P_{i-1} + \textit{Up}(\textit{Conv}(P_{i})), &i \in \{ 5,..,7 \}\\
        M = P_{i-1} + \textit{Up}(\textit{Conv}(P_{i})), & i=4.
    \end{cases}
    \label{con:fpnagg}
\end{equation}

\textbf{Scale-invariant Label Assignment}:
Existing label assignment algorithms heavily rely on accurate \textit{gt} scale information. For instance, FCOS assigns positive samples within the center region of \textit{gt} boxes, while Faster R-CNN designates positive samples when the IoU between anchors and \textit{gt} boxes exceeds 0.5. However, in the absence of \textit{gt} boxes, these assignment strategies cannot be applied. Therefore, we propose a one-to-one scale-invariant label assignment strategy. Specifically, we utilize the central points of the pseudo boxes generated by DMIL and select the nearest feature point as a positive sample based on the L1 distance to these central points.

\renewcommand{\arraystretch}{1.2}
\begin{table*}
	\centering
	\caption{Comparison between the Point Teacher and other methods on the AI-TOD-v2.0 {\tt val set} (HBB) and SODA-A~\citep{soda_2023_pami} {\tt val set} (OBB) with \textbf{central annotated points ($m$ = 0\%)}. We report $\rm{AP_{0.25}}$ in this table. Y denotes the end-to-end paradigm. We utilize FCOS as the detector except Point2Rbox-RC (YOLOF). * mean two-stage MIL.}
    \begin{tabular}{l | c | c | c c c c c c c c | c}
    \toprule
    Method & Type & E2E & VE & SH & PE/HE & ST & SP & AI & BR/CT & WM & mAP \\
    \midrule
    \multicolumn{12}{l}{\textit{HBox-supervised detectors (AI-TOD-v2)}} \\
    \cline{1-12}
    RetinaNet~\citep{Focal-Loss_2017_ICCV} & HBB & Y & 62.0 & 72.3 & 26.2 & 34.8 & 8.7 & 2.4 & 43.6 & 3.4 & 31.7 \\
    Faster R-CNN~\citep{faster-rcnn} & HBB & Y & 49.0 & 56.2 & 19.7 & 40.6 & 60.2 & 38.6 & 30.9 & 5.7 & 37.6 \\
    FCOS~\citep{fcos} & HBB & Y & 75.7 & 76.2 & 26.5 & 65.9 & 46.2 & 1.9 & 36.5 & 0.3 & 41.2 \\
    \cline{1-12}
    \multicolumn{12}{l}{\textit{Point-supervised detectors (AI-TOD-v2)}} \\
    \cline{1-12}
    P2BNet~\citep{p2bnet} & HBB & N & 0.9 & 1.9 & 0.0 & 7.6 & 0.1 & 0.0 & 1.2 & 0.0 & 1.5 \\
    P2BNet*~\citep{p2bnet} & HBB & N & 0.1 & 0.8 & 1.1 & 16.7 & 0.0 & 0.1 & 0.3 & 0.0 & 2.4 \\
    PLUG~\citep{plug} & HBB & N & 19.4 & 51.5 & 10.0 & 14.9 & 54.0 & 1.2 & 1.9 & 0.7 & 19.2 \\
    \cellcolor{gray!20}Point Teacher &
    \cellcolor{gray!20}HBB &
    \cellcolor{gray!20}Y & \cellcolor{gray!20}58.2 & \cellcolor{gray!20}62.8 & \cellcolor{gray!20}17.5 & \cellcolor{gray!20}47.1 & \cellcolor{gray!20}43.2 & \cellcolor{gray!20}1.3 & \cellcolor{gray!20}45.4 & \cellcolor{gray!20}8.7 & \cellcolor{gray!20}\textbf{35.5} \\
     \midrule
     \multicolumn{12}{l}{\textit{RBox-supervised detectors (SODA-A)}} \\
    \cline{1-12}
    RetinaNet-O~\citep{Focal-Loss_2017_ICCV} & OBB & Y & 50.1 & 82.2 & 53.0 & 77.2 & 88.4 & 89.0 & 58.8 & 87.0 & 70.7 \\
    FCOS-O~\citep{fcos} & OBB & Y & 61.9 & 87.4 & 51.5 & 81.6 & 88.6 & 89.8 & 60.2 & 88.7 & 74.6 \\
    Oriented R-CNN~\citep{orcnn} & OBB & Y & 61.0 & 87.1 & 56.0 & 84.9 & 88.6 & 89.6 & 55.5 & 88.9 & 74.7 \\
    \cline{1-12}
    \multicolumn{12}{l}{\textit{Point-supervised detectors (SODA-A)}} \\
    \cline{1-12}
    PointOBB*~\citep{pointobb} & OBB & N & 11.7 & 13.7 & 56.0 & 19.7 & 45.3 & 80.4 & 22.8 & 78.7 & 37.8 \\
    Point2Rbox-RC~\citep{point2rbox} & OBB & Y & 6.3 & 14.4 & 71.1 & 48.4 & 88.8 & 10.1 & 3.3 & 76.0 & 36.1 \\
    PointOBB-v2~\citep{pointobbv2} & OBB & N & 26.3 & 66.4 & 50.5 & 55.0 & 41.4 & 56.0 & 44.6 & 7.2 & 41.5 \\
    \cellcolor{gray!20}Point Teacher &
    \cellcolor{gray!20}OBB &
    \cellcolor{gray!20}Y & \cellcolor{gray!20}43.9 & \cellcolor{gray!20}77.6 & \cellcolor{gray!20}27.5 & \cellcolor{gray!20}22.5 & \cellcolor{gray!20}85.8 & \cellcolor{gray!20}31.8 & \cellcolor{gray!20}19.0 & \cellcolor{gray!20}72.9 & \cellcolor{gray!20}\textbf{47.2} \\
    \bottomrule
    \end{tabular}
	\label{tab:main_results}
\end{table*}

\section{Experiments}
\label{experiments}
\subsection{Experimental Settings}
\label{implementationDetails}
\textbf{Point-annotated Dataset.} 
We conduct comprehensive evaluations of our method on the AI-TOD-v2.0~\citep{aitodv2_2022_isprs} dataset, a benchmark known for its challenging tiny object detection scenarios with an average object size of 12.7 pixels. Additionally, we validate the effectiveness of our method in small aerial object detection scenarios using the TinyPerson~\citep{TinyPerson_2020_WACV} and SODA~\citep{soda_2023_pami} datasets.
Based on these datasets, we propose a method for generating point annotations. The location of the point is defined within a range $m$ around the region of the \textit{gt}, where $m$ varies from 0\% to 100\%. 
Specifically, let $\{cx, cy, w, h \}$ and $\{cx, cy, w, h, \theta \}$ represent an object in the HBB and OBB tasks, respectively. We simulate an annotated point $\{px, py\}$ as follows:
\begin{equation}
    \begin{cases}
        px_{h} = cx + \Delta_x \cdot w\\ 
        py_{h} = cy + \Delta_y \cdot h\\
        px_{o} = cx + \Delta_x \cdot w \cdot cos\theta\\ py_{o} = cy + \Delta_y \cdot h \cdot sin\theta,
    \end{cases}
    \label{con:hbbsimulate}
\end{equation}
where $\Delta_x$ and $\Delta_y$ follow the uniform distribution $U(-m/2,m/2)$. 
When $m = 0\%$, the point is placed at the center of the \textit{gt}, while $m = 100\%$ allows the point to be located anywhere within the \textit{gt}. Notably, to validate the robustness of Point Teacher, we conduct main experiments on AI-TOD-v2.0 and SODA-A under varying conditions with $m$ = 0\% to 100\%. Ablation studies are performed specifically with $m$ = 0\% to ablate the effects of each component.

\textbf{Implementation Details.} 
Our implementation is based on the MMDetection~\citep{mmdetection} and MMRotate~\citep{mmrotate} toolkits, built on the PyTorch~\citep{PyTorch_2019_NIPS} deep learning framework. We employ an ImageNet~\citep{ImageNet_2015_IJCV} pre-trained model as the backbone. Training is conducted for 12 epochs using the Stochastic Gradient Descent (SGD) optimizer, with a momentum of 0.9, a weight decay of 0.0001, and a batch size of 2. The initial learning rate is set to 0.005 and decreases at epochs 8 and 11. The Region Proposal Network (RPN) generates up to 3000 proposals. During inference, we filter background boxes with a confidence threshold of 0.05 and apply Non-Maximum Suppression (NMS) with an IoU threshold of 0.5, selecting the top 3000 bounding boxes. All other parameters remain consistent with the defaults in MMDetection and MMRotate.
% The evaluation metric includes AP$_{0.25}$, AP$_{vt}$, AP$_{t}$, AP$_{s}$, and AP$_{m}$. 
Given that IoU-based metrics are particularly unfavorable for tiny object detection~\citep{aitodv2_2022_isprs}, AP$_{0.5}$ is not an ideal evaluation standard under point annotation scenarios. Therefore, we adopt AP$_{0.25}$ as an alternative evaluation metric in this work.
The teacher model is an exponential moving average (EMA) of the student model, with the EMA momentum set to the default value of 0.999~\citep{softteacher}. The fusion weight $\beta$ for generating pseudo boxes is set to 0.25, while the shaking ratio $r$ for Jittering IoU Loss is chosen to be 0.2. During fusion, $K_1$, $K_2$, and $K_3$ are set to 5, 3, and 1, respectively.
The Spatial-aware Box Generation phase occurs during the first 4000 iterations of training, with the remaining iterations dedicated to the Noise-aware Label Evolution phase.

\renewcommand{\arraystretch}{1.2}
\begin{table*}
	\centering
	\caption{Comparison between other methods and our proposed Point Teacher on the AI-TOD-v2.0 {\tt val set} (HBB) and SODA-A~\citep{soda_2023_pami} {\tt val set} (OBB) with \textbf{randomly annotated points ($m$ = 100\%)}.}
    \begin{tabular}{l | c | c | c c c c c c c c | c}
    \toprule
    Method & Type & E2E & VE & SH & PE/HE & ST & SP & AI & BR/CT & WM & mAP \\
    \midrule
    \multicolumn{12}{l}{\textit{HBox-supervised detectors (AI-TOD-v2)}} \\
    \cline{1-12}
    RetinaNet~\citep{Focal-Loss_2017_ICCV} & HBB & Y & 62.0 & 72.3 & 26.2 & 34.8 & 8.7 & 2.4 & 43.6 & 3.4 & 31.7 \\
    Faster R-CNN~\citep{faster-rcnn} & HBB & Y & 49.0 & 56.2 & 19.7 & 40.6 & 60.2 & 38.6 & 30.9 & 5.7 & 37.6 \\
    FCOS~\citep{fcos} & HBB & Y & 75.7 & 76.2 & 26.5 & 65.9 & 46.2 & 1.9 & 36.5 & 0.3 & 41.2 \\
    \cline{1-12}
    \multicolumn{12}{l}{\textit{Point-supervised detectors (AI-TOD-v2)}} \\
    \cline{1-12}
    P2BNet~\citep{p2bnet} & HBB & N & 0.2 & 0.5 & 0.0 & 1.8 & 0.0 & 0.0 & 0.5 & 0.0 & 0.4 \\
    P2BNet*~\citep{p2bnet} & HBB & N & 0.6 & 0.7 & 0.8 & 5.5 & 0.0 & 0.1 & 3.4 & 0.0 & 1.4 \\
    PLUG~\citep{plug} & HBB & N & 3.8 & 2.3 & 2.0 & 4.1 & 48.3 & 3.0 & 3.0 & 0.0 & 8.3 \\
    \cellcolor{gray!20}Point Teacher &
    \cellcolor{gray!20}HBB &
    \cellcolor{gray!20}Y & \cellcolor{gray!20}54.5 & \cellcolor{gray!20}63.1 & \cellcolor{gray!20}16.2 & \cellcolor{gray!20}40.4 & \cellcolor{gray!20}33.9 & \cellcolor{gray!20}0.7 & \cellcolor{gray!20}41.9 & \cellcolor{gray!20}2.3 & \cellcolor{gray!20}\textbf{31.6} \\
     \midrule
     \multicolumn{12}{l}{\textit{RBox-supervised detectors (SODA-A)}} \\
    \cline{1-12}
    RetinaNet-O~\citep{Focal-Loss_2017_ICCV} & OBB & Y & 50.1 & 82.2 & 53.0 & 77.2 & 88.4 & 89.0 & 58.8 & 87.0 & 70.7 \\
    FCOS-O~\citep{fcos} & OBB & Y & 61.9 & 87.4 & 51.5 & 81.6 & 88.6 & 89.8 & 60.2 & 88.7 & 74.6 \\
    Oriented R-CNN~\citep{orcnn} & OBB & Y & 61.0 & 87.1 & 56.0 & 84.9 & 88.6 & 89.6 & 55.5 & 88.9 & 74.7 \\
    \cline{1-12}
    \multicolumn{12}{l}{\textit{Point-supervised detectors (SODA-A)}} \\
    \cline{1-12}
    PointOBB*~\citep{pointobb} & OBB & N & 3.3 & 2.2 & 49.4 & 5.4 & 65.8 & 48.6 & 0.5 & 72.6 & 27.9 \\
    Point2Rbox-RC~\citep{point2rbox} & OBB & Y & 0.0 & 0.5 & 0.1 & 0.1 & 7.3 & 0.0 & 0.0 & 1.3 & 1.0 \\
    PointOBB-v2~\citep{pointobbv2} & OBB & N & 10.1 & 42.8 & 22.5 & 43.9 & 12.4 & 56.0 & 8.0 & 2.2 & 23.1 \\
    \cellcolor{gray!20}Point Teacher &
    \cellcolor{gray!20}OBB &
    \cellcolor{gray!20}Y & \cellcolor{gray!20}16.6 & \cellcolor{gray!20}34.2 & \cellcolor{gray!20}42.2 & \cellcolor{gray!20}62.7 & \cellcolor{gray!20}43.8 & \cellcolor{gray!20}71.3 & \cellcolor{gray!20}13.0 & \cellcolor{gray!20}59.3 & \cellcolor{gray!20}\textbf{40.0} \\
    \bottomrule
    \end{tabular}
	\label{tab:main_results_random}
\end{table*}

% \renewcommand{\arraystretch}{1.5}
% \begin{table}
% 	\centering
% 	\caption{Performance of the Point Teacher with the Point Position $m$ set to 0\%, 30\%, 60\%, and 100\%.}
%     % \resizebox{\linewidth}{!}{
% 	\begin{tabular}{c | c c c c c}
%         \toprule
%         $m$ & mAP & AP$_{vt}$ & AP$_{t}$ & AP$_{s}$ & AP$_{m}$ \\
%         \cline{1-6}
%         0\% & 35.5 & 17.5 & 42.0 & 37.4 & 26.6 \\
%         % \cline{1-6}
%         30\% & \textbf{36.3} & 21.2 & 44.5 & 32.0 & 38.2 \\
%         % \cline{1-6}
%         60\% & 33.0 & 19.3 & 41.5 & 32.1 & 36.1 \\
%         % \cline{1-6}
%         100\% & 31.3 & 12.8 & 46.8 & 27.6 & 36.5 \\
%         \bottomrule
%         \end{tabular}
% 	\label{tab:robustness}
% \end{table}

\renewcommand{\arraystretch}{1.3}
\begin{table}
	\centering
	\caption{ Robustness of other methods and our proposed Point Teacher performance with the point location $m$ set to 0\%, 30\%, 60\%, and 100\%. * means two-stage MIL.}
    % \resizebox{\linewidth}{!}{
	\begin{tabular}{c | c c c c}
        \toprule
        $m$ & Point Teacher & PLUG & P2BNet* & P2BNet  \\
        \cline{1-5}
        0\% & 35.5 & 19.2 & 2.4 & 1.5  \\
        30\% & \textbf{36.3} & 16.7 & 5.7 & 1.0  \\
        60\% & 33.0 & 13.5 & 3.4 & 0.6  \\
        100\% & 31.6 & 8.3 & 1.4 & 0.4  \\
        \bottomrule
        \end{tabular}
	\label{tab:robustness}
\end{table}

\renewcommand{\arraystretch}{1.3}
\begin{table}
	\centering
	\caption{Comparison between other methods and our proposed Point Teacher on the TinyPerson~\citep{TinyPerson_2020_WACV} {\tt test set} with central annotated points. We report $\rm{AP_{0.25}}$ in this table. We utilize FCOS as the detector.}
    % \resizebox{\linewidth}{!}{
	\begin{tabular}{l | c c c c c}
        \toprule
        Method & mAP & AP$_{vt}$ & AP$_{t}$ & AP$_{s}$ & AP$_{m}$ \\
        \cline{1-6}
        \multicolumn{6}{l}{\textit{Hbox-supervised detectors}} \\
        \cline{1-6}
        RetinaNet & 13.9 & 2.3 & 7.0 & 16.1 & 33.9 \\
        Faster R-CNN & 15.6 & 0.0 & 3.8 & 19.2 & 52.2 \\
        FCOS & 34.2 & 9.1 & 24.2 & 43.5 & 68.8 \\
        \cline{1-6}
        \multicolumn{6}{l}{\textit{Point-supervised detectors}} \\
        \cline{1-6}
        P2BNet  & 2.3 & 0.0 & 1.0 & 4.0 & 8.3 \\
        P2BNet* & 7.9 & 0.0 & 2.4 & 14.1 & 24.2 \\
        PLUG    & 7.4 & 0.0 & 1.0 & 8.9 & 28.1 \\
        \rowcolor{gray!20}Ours & \textbf{18.6} & 3.8 & 19.3 & 24.4 & 29.4 \\
        \bottomrule
        \end{tabular}
	\label{tab:tinyperson}
\end{table}

\subsection{Main Results}
We compare our method with state-of-the-art (SOTA) methods on the AI-TOD-v2.0~\citep{aitodv2_2022_isprs} (horizontal object detection) and SODA-A~\citep{soda_2023_pami} (oriented object detection) datasets. As shown in Table~\ref{tab:main_results}, our method consistently outperforms all current SOTA algorithms on tiny object detection tasks. For the horizontal object detection task, we compare our approach with P2BNet~\citep{p2bnet} and PLUG~\citep{plug}, where P2BNet exhibits a weak performance of 2.4\%. This is mainly because P2BNet, a MIL-based pseudo-boxes generator, relies solely on classification scores to filter pseudo boxes, but the weak features of tiny objects make the classification scores inaccurate in reflecting the quality of pseudo boxes. PLUG, a CPM-based pseudo-boxes generator, produces relatively accurate pseudo-boxes. However, due to the indistinct boundaries and color features of tiny objects, the segmentation loss in PLUG fails to converge effectively. In contrast, our method, after obtaining coarse pseudo boxes during the Spatial-aware Box Generation phase, leverages the spatially-aware DMIL module to generate more stable and accurate pseudo boxes, thereby improving detection performance by 16.3\%.
For the oriented object detection task, we compare our approach with PointOBB~\citep{pointobb}, PointOBB-v2~\citep{pointobbv2} and Point2Rbox~\citep{point2rbox}. Although PointOBB remains a MIL-based method, the issue of classification scores is mitigated due to the relatively larger object sizes in the SODA-A dataset, resulting in a significant improvement in accuracy of 37.8\%. Point2Rbox, as an Auxiliary-based method, also demonstrates strong competitiveness. However, it is important to note that the indistinct features of tiny objects create a domain gap between the synthetic objects and the real tiny objects, limiting their generalization ability. During training, the network tends to overfit the features of the synthetic objects, which deteriorates the regression branch. In contrast, our method alleviates this issue by utilizing pseudo boxes generated by DMIL for supervision, leading to an improvement in performance by 11.1\%.
Unlike the significant improvement observed in the HBB task, the performance of the Point Teacher on the OBB task is relatively lower. This is primarily because our method is directly transferred from HBB to OBB without any refinement or dedicated design to address angle-related issues.

We also conduct experiments on the TinyPerson~\citep{TinyPerson_2020_WACV} dataset. The results, as shown in Table~\ref{tab:tinyperson}, demonstrate that our method achieves competitive performance, reaching 54.4\% of hbox-supervised accuracy.

\subsection{Robustness of Point Location}
To assess the robustness of our method to variations in point location, we conduct a series of detailed experiments. First, we perform the primary evaluation on the AI-TOD-v2.0 and SODA-A datasets under fully randomized point location setting ($m$ = 100\%), as shown in Table~\ref{tab:main_results_random}. Compared with the results for the center-point setting in Table~\ref{tab:main_results}, all methods experienced some performance degradation. Point2Rbox exhibits the most significant decrease by 35.1\%, as it relies on center points as priors in its label assignment strategy, which leads to substantial performance drops under randomized points. Similarly, MIL-based (\textit{e.g.,} P2BNet, PointOBB) and CPM-based (\textit{e.g.,} PLUG, PointOBB-v2) methods also show some decline under this condition. In contrast, our proposed denoising-based method demonstrated strong robustness, with performance decreasing by only 3.9\% and 7.3\% on the AI-TOD-v2.0 and SODA-A datasets, respectively. 
Additionally, we further analyze results for different values of the point location parameter $m$ set to 0\%, 30\%, 60\%, and 100\%. The results, shown in Table~\ref{tab:robustness}, indicate that our method achieves consistently high accuracy regardless of the point location. Instead, other methods such as P2BNet and PLUG experience a more significant drop in accuracy as the point location varies. Notably, when the point is placed at the center (\textit{i.e.}, $m=0\%$), the accuracy is slightly lower than at $m=30\%$ by 0.8\%. This is because placing the point at the center provides prior information, leading the network to learn a central bias. In contrast, setting the point location at 30\% enhances the network's robustness, resulting in more accurate predictions. 

\begin{table*}[t]\vspace{-3mm}
% subfloat a
\centering
\caption{Ablations. We train on point-based AI-TOD-v2.0 \texttt{train set}, test on \texttt{val set}. 
Phase1 and Phase2 represent Spatial-aware Box Generation phase and Noise-aware Label Evolution phase respectively.
Gray row means default setting.}
\subfloat[Individual effectiveness of components in Point Teacher. \label{tab:ablation}]{
\tablestyle{2pt}{1.2}\begin{tabular}{c | c c c | c}
	\toprule
    Phase1 & \multicolumn{3}{c|}{Phase2} & \multirow{2}{*}{mAP} \\
    \cline{1-4}
    Randomly Mask & Teacher-Student & DMIL & Jittering IoU Loss & \\
    \cline{1-5}
    \checkmark & & & & 21.0 \\
    \checkmark & \checkmark & & & 24.5 \\
    \checkmark & \checkmark & \checkmark & & 34.7 \\
    \rowcolor{gray!20} \checkmark & \checkmark & \checkmark & \checkmark & \textbf{35.5} \\
    \bottomrule
\end{tabular}}\hspace{3mm}
% subfloat b
\subfloat[Comparison of DMIL and MIL. * denotes two-stage MIL. \label{tab:ablation_mil}]{
\tablestyle{10pt}{1.8}\begin{tabular}{c | c c c c c}  
	\toprule
     Method & mAP & AP$_{vt}$ & AP$_{t}$ & AP$_{s}$ & AP$_{m}$ \\
    \cline{1-6}
    MIL  & 20.0 & 10.0 & 27.6 & 18.3 & 17.4 \\
    MIL* & 20.3 & 9.7 & 29.6 & 15.7 & 22.9 \\
    \cline{1-6}
    \rowcolor{gray!20} DMIL & \textbf{35.5} & 17.5 & 42.0 & 37.4 & 26.6 \\
    \bottomrule
\end{tabular}}\\
% subfloat c
\subfloat[\{$K_1$, $K_2$, $K_3$\} for Pseudo Boxes Generation.
\label{tab:ablation_k1k2k3}]{
\tablestyle{8pt}{1.4}\begin{tabular}{c c c | c c c c c}  
	\toprule
    $K_1$ & $K_2$ & $K_3$ & mAP & AP$_{vt}$ & AP$_{t}$ & AP$_{s}$ & AP$_{m}$ \\
    \cline{1-8}
    3 & 1 & 1 & 28.2 & 25.3 & 31.8 & 42.7 & 0.4 \\
    3 & 1 & 5 & 33.6 & 28.2 & 37.1 & 44.3 & 14.4 \\
    \rowcolor{gray!20} 5 & 3 & 1 & \textbf{35.5} & 17.5 & 42.0 & 37.4 & 26.6 \\
    5 & 3 & 5 & 31.9 & 10.3 & 47.3 & 30.1 & 55.5 \\
    \bottomrule
\end{tabular}}\hspace{3mm}
% subfloat d
\subfloat[$r$ in Jittering IoU Loss. 
\label{tab:ablation_r}]{
\tablestyle{12pt}{1.4}\begin{tabular}{c | c c c c c}  
	\toprule
    $r$ & mAP & AP$_{vt}$ & AP$_{t}$ & AP$_{s}$ & AP$_{m}$ \\
    \cline{1-6}
    0.0 & 34.7 & 12.4 & 44.5 & 32.0 & 21.9 \\
    \rowcolor{gray!20} 0.2 & \textbf{35.5} & 17.5 & 42.0 & 37.4 & 26.6 \\
    0.4 & 33.2 & 6.3 & 56.0 & 30.2 & 57.0 \\
    0.6 & 33.0 & 5.6 & 44.1 & 37.6 & 49.1 \\
    \bottomrule
\end{tabular}}\\
% subfloat e
\subfloat[$\beta$ in DMIL Fusion.
\label{tab:ablation_beta}]{
\tablestyle{12pt}{1.45}\begin{tabular}{c | c c c c c}  
	\toprule
    $\beta$ & mAP & AP$_{vt}$ & AP$_{t}$ & AP$_{s}$ & AP$_{m}$ \\
    \cline{1-6}
    0.00 & 34.7 & 22.6 & 40.9 & 46.6 & 10.9 \\
    \rowcolor{gray!20} 0.25 & \textbf{35.5} & 17.5 & 42.0 & 37.4 & 26.6 \\
    0.50 & 31.6 & 13.9 & 44.0 & 29.9 & 58.8 \\
    1.00 & 24.5 & 5.4 & 38.8 & 23.1 & 32.1 \\
    \bottomrule
\end{tabular}}\hspace{3mm}
% subfloat f
\subfloat[Training proportion of the Spatial-aware Box Generation phase.
\label{tab:ablation_phase1}]{
\tablestyle{6pt}{1.45}\begin{tabular}{c c | c c c c c}  
	\toprule
    Iterations & Proportion & mAP & AP$_{vt}$ & AP$_{t}$ & AP$_{s}$ & AP$_{m}$ \\
    \cline{1-7}
    \rowcolor{gray!20} 4000 & 5\% & \textbf{35.5} & 17.5 & 42.0 & 37.4 & 26.6 \\
    20000 & 25\% & 34.2 & 5.3 & 53.4 & 30.9 & 51.8 \\
    40000 & 50\% & 34.1 & 7.9 & 50.1 & 34.0 & 52.2 \\
    80000 & 100\% & 21.0 & 8.0 & 31.4 & 18.8 & 21.9 \\
    \bottomrule
\end{tabular}}
% main caption
\label{tab:ablations}\vspace{-3mm}
\end{table*}

\subsection{Ablation Study}
This section investigates the contribution of key designs
and the selection of hyper-parameters. First, we verify the
effects of each module in Point Teacher. Then, we study the
influence of the selection of different hyper-parameters. Note
that ablations are performed on the AI-TOD-v2.0 dataset with central annotated points.

\textbf{Modules in Point Teacher:} To further validate the effectiveness of each module in our proposed method, we conduct ablation experiments on the AI-TOD-v2.0 dataset, with the results shown in Table~\ref{tab:ablation}.
%Specifically, our approach consists of a Spatial-aware Box Generation phase and a Noise-aware Label Evolution phase. 
When only the Spatial-aware Box Generation phase is utilized, the network initially develops spatial awareness. As a result, the accuracy remains relatively low, achieving only 21.0 AP. When the Noise-aware Label Evolution phase is introduced, the network’s performance improves. The introduction of pseudo boxes by the teacher network provides supervision that improves overall performance. However, due to the coarse nature of these pseudo boxes, the accuracy increases by only 3.5\%. With the addition of the DMIL module, the coarse pseudo boxes are refined, providing more stable and accurate supervision signals. Finally, with the integration of Jittering IoU Loss, the network’s resistance to noisy bounding boxes is significantly enhanced, further boosting overall performance by 11.0\%.

Moreover, to validate that our proposed DMIL provides more accurate supervision than MIL, we replace the DMIL with one-stage MIL and two-stage MIL*~\citep{p2bnet}. As shown in Table~\ref{tab:ablation_mil}, our method outperforms existing MIL approaches, achieving the highest accuracy.

\textbf{\{$K_1$, $K_2$, $K_3$\} for Pseudo Boxes Generation:} 
$K_1$, $K_2$, and $K_3$ are hyperparameters that control the generation of pseudo boxes. Specifically, $K_1$ and $K_2$ work in tandem to guide the teacher network in producing coarse pseudo boxes, while $K_3$ is used in the DMIL module for Instance Selection to filter high-quality proposals. As shown in Table~\ref{tab:ablation_k1k2k3}, when $K_1$ and $K_2$ are set to {5, 3} for generating coarse pseudo boxes, the results significantly outperform those obtained with {3, 1}. This is because a greater number of proposals increases the likelihood of generating more accurate boxes and aids in constructing proposals for medium-sized objects. For $K_3$, a value of 1 proves more effective for small objects compared to 5, as using more proposals to refine pseudo boxes tends to result in expanded boxes, which is better suited for medium-sized objects.

\textbf{$r$ in Jittering IoU Loss:} 
$r$ controls the degree of perturbation in the Jittering IoU Loss. When $r$ is set to 0, Jittering IoU Loss degrades to the standard IoU Loss. As shown in Table~\ref{tab:ablation_r}, the best performance is achieved with $r$ set to 0.2, while increasing $r$ to 0.4 and 0.6 leads to a drop in accuracy. This is because a slight level of perturbation introduces a beneficial noise term in the regression process, preventing overfitting to inaccurate regression targets. However, as $r$ increases, the perturbations become too pronounced, causing the network to learn incorrect information and resulting in a decline in accuracy.

\begin{figure*}[t]
    \centering
    \includegraphics[width=0.98\linewidth]{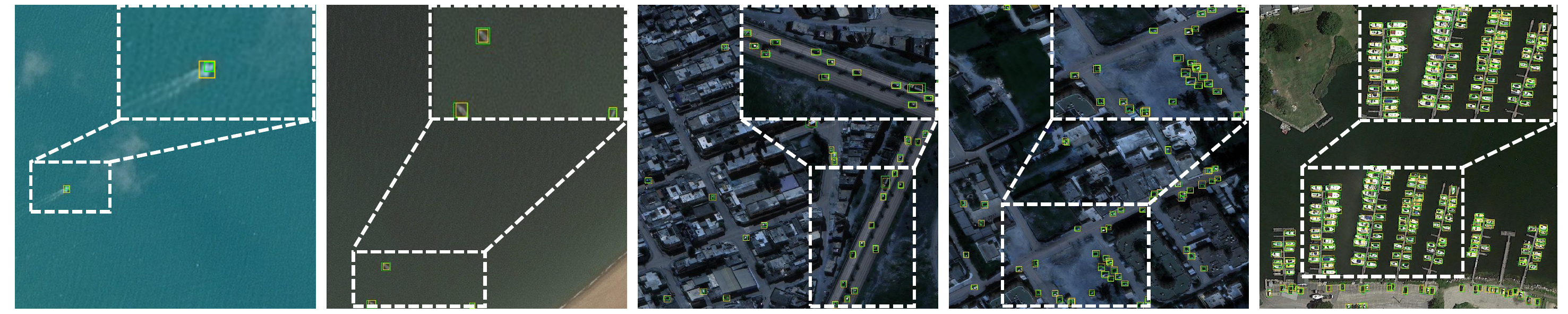}
    \caption{Visualization of pseudo boxes generated by the DMIL Module. Green boxes denote the \textit{gt} boxes, yellow boxes denote the pseudo boxes generated by DMIL.}
    \label{fig:pseudoboxes}
    %\vspace{\fixedvskip}
\end{figure*}

\textbf{$\beta$ in DMIL Fusion:} 
$\beta$ controls the fusion weighting between coarse pseudo boxes and the proposals selected through DMIL. $\beta$ is a widely used hyperparameter in generating pseudo boxes~\citep{OAMIL, dntod}. When $\beta$ is set to 0, only the proposals filtered by DMIL are used to generate refined pseudo boxes. Conversely, when $\beta$ is set to 1, only the coarse pseudo boxes are used for supervision. As shown in Table~\ref{tab:ablation_beta}, setting $\beta$ to 0 can lead to object drift (\textit{i.e.}, pseudo boxes being predicted over higher-confidence objects in dense scenarios), resulting in reduced accuracy. On the other hand, relying solely on coarse pseudo boxes with $\beta$ set to 1 yields inaccurate predictions. The optimal performance is achieved by balancing the two, as combining both sources of supervision provides a more reliable outcome. 
% Interestingly, the model is not highly sensitive to $\beta$ values, though generally, when the coarse pseudo boxes are sufficiently accurate, a larger $\beta$ tends to give better results.

\textbf{Training Time of the First Phase:} 
The Spatial-aware Box Generation phase (\textit{i.e.,} Phase1) occurs during the first 4000 iterations (5\% of the total training iterations) to provide the network with initial spatial awareness. To evaluate the impact of the duration of this phase on the overall network performance, we conduct experiments with varying Phase1 durations of 5\%, 25\%, 50\%, and 100\% of the total training iterations. The experimental results, as shown in Table~\ref{tab:ablation_phase1}, reveal that as the proportion of Phase1 increases, network accuracy gradually declines. This behavior can be attributed to the mask-guided learning approach, which only enables the network to acquire coarse spatial awareness. This assistance does not scale with training duration, and excessively prolonged initialization reduces the time allocated for the denoising learning phase (\textit{i.e.,} Phase2), thus deteriorating overall network performance. Notably, when Phase1 constitutes 100\% of the training (with Phase2 excluded), the accuracy is significantly lower compared to when both phases are employed.

\subsection{Visual Analysis}
We conduct a series of analytical experiments to demonstrate that our methods can provide reliable box supervision for training. First, we visualize the pseudo boxes generated by the DMIL module and the \textit{gt} boxes, as shown in Figure~\ref{fig:pseudoboxes}. The pseudo boxes generated by DMIL are more accurate and closely align with the \textit{gt} boxes. However, in densely arranged scenes (as illustrated in the fourth column of the figure), due to DMIL's reliance on classification scores during the refinement process, instances of overlapping predictions occur.
Second, we visualize the detection results of our method on the test set and compare them with the SOTA algorithm PLUG. The results are presented in Figure~\ref{fig:compare}, where it can be observed that, even with only point annotations, our method can produce reasonably accurate predictions in an end-to-end training scenario.

\section{Discussion}
\label{discussion}
In this work, we explore the potential of detecting tiny objects with low-cost annotations, specifically leveraging point annotations. Observing the limited scale and ambiguous boundaries make it challenging to annotate accurately on the main body of the tiny object, we propose a point-robust method, Point Teacher, to address these issues.

To further advance the field of point-supervised tiny object detection, this section focuses on addressing the following three critical questions:

\begin{figure*}[t]
    \centering
    \includegraphics[width=0.98\linewidth]{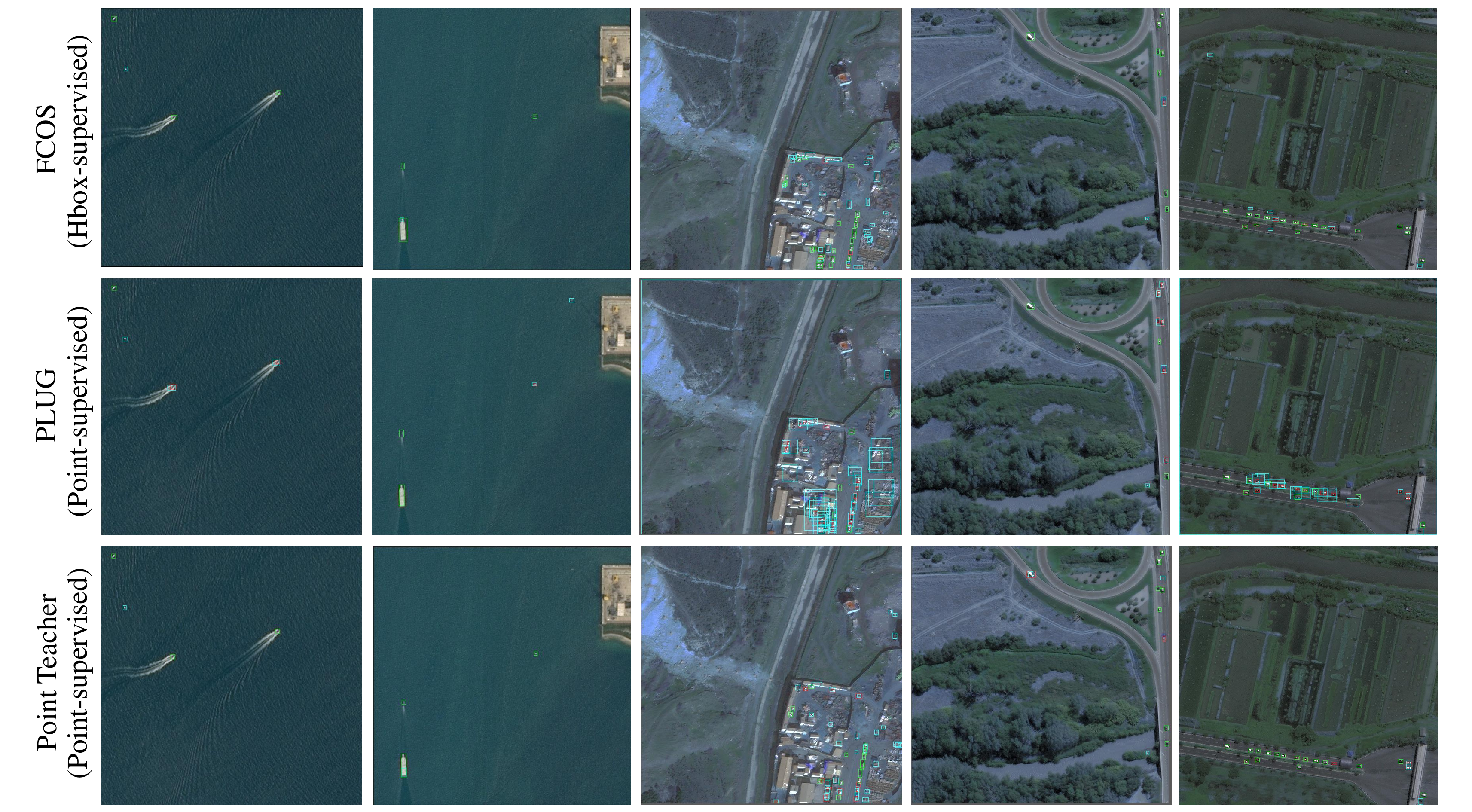}
    \caption{Visualization results on AI-TOD-v2.0 with central annotated points. The first row is the result of FCOS, the second row is the result of PLUG, and the third row is the result of Point Teacher. Green boxes denote true positive predictions, red boxes denote false negative predictions, and blue boxes denote false positive predictions.}
    \label{fig:compare}
    %\vspace{\fixedvskip}
\end{figure*}

-- \textit{Why study point supervision instead of other label-efficient methods for tiny object detection?}

Point annotations are not only well-suited to the unique characteristics of tiny objects but also strike a balance between annotation cost and detection accuracy. Due to the small scale (less than $16 \times 16$ pixels), tiny objects exhibit point-like distributions in feature maps. Compared to larger objects, point annotations provide stronger prior information, making them especially advantageous for tiny object detection. Furthermore, while point annotations are slightly more costly than image-level annotations~\citep{p2bnet}, they achieve accuracy comparable to box-level annotations, offering a practical trade-off.
In contrast, annotation types such as scribbles or other detailed forms introduce higher costs and often include excessive information that is redundant for tiny object detection tasks~\citep{ufo2}. Point annotations, by contrast, are precise and efficient, making them an optimal choice for annotating tiny objects.

-- \textit{Why do existing point-supervised methods perform poorly on tiny object detection?}

The weak feature representation and ambiguous boundaries of tiny objects present significant challenges for adapting existing methods to this domain. Current approaches primarily fall into three categories: MIL-based~\citep{p2bnet, pointobb}, CPM-based~\citep{plug, pointobbv2}, and Auxiliary-based~\citep{point2rbox, p2rbox} methods.
MIL-based methods rely on classification scores to select proposals within a bag as supervision. However, the weak features of tiny objects make it difficult to distinguish between proposals based on classification scores, limiting the applicability of MIL-based methods to tiny object detection.
CPM-based methods generate pseudo boxes using class probability map (CPM). However, the small scale and ambiguous boundaries of tiny objects undermine the boundary saliency in CPM, reducing the effectiveness of these methods.
For Auxiliary-based methods (excluding SAM-based approaches), the weak feature representation of tiny objects impedes the generalization of regression training with auxiliary information, hindering further improvements in detection accuracy.
In contrast, our proposed Point Teacher introduces a two-phase denoising-based paradigm, progressively refining the quality of pseudo boxes at each phase. This approach ensures more robust supervision, effectively addressing the challenges posed by tiny object detection.

-- \textit{What are the limitations of Point Teacher?}

Despite the significant advancements achieved by Point Teacher, there are remaining challenges. 
First, while our method excels at point-supervised tiny object detection, its performance on multi-scale objects could be further optimized. This limitation arises because the Spatial-aware Box Generation phase relies on randomly masked regions to enhance spatial awareness. However, the masks used are simple, with limited shape and color variation, which restricts the model’s ability to generalize across multi-scale objects. To address this, future work could incorporate more diverse mask patterns or adopt multi-scale features in the spatial-aware phase to improve the model’s generalization. 
Second, the challenge of accurately detecting densely packed tiny objects is further exacerbated by the position noise inherent in point annotations, which significantly hinders precise localization. Although our method leverages spatial-aware DMIL for improved localization guidance, multiple instance learning still struggles to effectively filter pseudo boxes in densely arranged object scenarios. Future work could explore the use of vision-language models~\citep{clip, sam}, which offer enhanced text-visual alignment capabilities, to provide more robust guidance in dense object arrangements.

% conclusion
\section{Conclusion}
\label{conclusion}
Single point supervision offers a cost-effective solution for labeling large-scale tiny object datasets. However, the inherent challenges of tiny objects—such as their small size and weak features—make them highly sensitive to the precision of point locations. In this paper, we investigate the robustness of point-supervised tiny object detection under varying point locations and introduce Point Teacher, a robust end-to-end point-based detector.
Point Teacher consists of a two-phase denoising-based learning paradigm specifically aimed at mitigating point localization noise. In the Spatial-aware Box Generation phase, spatial awareness is enhanced by randomly masked regions of the image, prompting the network to better predict spatial patterns. The Noise-aware Label Evolution phase employs a teacher-student architecture with DMIL to refine pseudo boxes and improve detection accuracy. Moreover, we introduce a novel Jittering IoU loss to prevent the model from overfitting to noisy pseudo boxes, thereby further enhancing robustness. Extensive experiments conducted on tiny object datasets demonstrate that Point Teacher surpasses existing methods, offering superior robustness and accuracy in point-supervised tiny object detection.

\section*{Acknowledgements}
The research was partially supported by the National Natural Science Foundation of China (NSFC) under Grants 62271355. The numerical calculations were conducted on the supercomputing system in the Supercomputing Center, Wuhan University. 

\printcredits

\bibliography{Jinwang-Papers, chang, zhuhaoran}
%\printbibliography
\end{sloppypar}
\end{document}